\def\paperID{38565}
\def\confName{CVPR}
\def\confYear{2026}

\def\paperTitle{Eulerian Gaussian Splatting using Hashed Probability Pyramids}

\def\authorBlock{
    Mia Gaia Polansky$^{1}$ \qquad
    George Kopanas$^{2}$ \qquad \\
    Stephan Garbin$^{3}$ \qquad
    Todd Zickler$^{1}$ \qquad
    Dor Verbin$^{2}$ \\
    [0.75em]
    $^{1}$Harvard University \qquad
    $^{2}$Google DeepMind
    \qquad
    $^{3}$Google\\
}

\newif\ifreview 
\newif\ifarxiv \newcommand{\arxiv}{\arxivtrue}
\newif\ifcamera 
\newif\ifrebuttal 
\arxiv

\pdfoutput=1
\documentclass[10pt,twocolumn,letterpaper]{article}
\ifreview \usepackage[review]{cvpr} \fi
\ifarxiv \usepackage[pagenumbers]{cvpr} \fi
\ifrebuttal \usepackage[rebuttal]{cvpr} \fi
\ifcamera \usepackage{cvpr} \fi

\usepackage{graphicx}	
\usepackage{amsmath}	
\usepackage{amssymb}	
\usepackage{booktabs}
\usepackage{times}
\usepackage{microtype}
\usepackage{epsfig}
\usepackage{caption}
\usepackage{float}
\usepackage{placeins}
\usepackage{color, colortbl}
\usepackage{stfloats}
\usepackage{enumitem}
\usepackage{tabularx}
\usepackage{xstring}
\usepackage{multirow}
\usepackage{xspace}
\usepackage{url}
\usepackage{subcaption}
\usepackage{xcolor}
\usepackage{tikz}
\usepackage{accents}
\usepackage[hang,flushmargin]{footmisc}

\usepackage{algorithm}
\usepackage{algpseudocode}

\ifcamera \usepackage[accsupp]{axessibility} \fi

\usepackage{listings}
\usepackage{soul}
\usepackage{subfiles}
\usepackage{tikz}
\usepackage{mathtools}
\usepackage{booktabs,multirow,makecell,tabularx,array,graphicx}
\usepackage{nicematrix}
\usepackage{nicefrac}

\usepackage{cancel}

% ...

\DeclareMathOperator{\volindex}{\nu}

\DeclareMathOperator{\bmu}{\boldsymbol{\mu}}

\newcommand{\p}{p}
\newcommand{\half}{\nicefrac{1}{2}}
\newcommand{\IGT}{I^{\mathrm{GT}}}

\newcommand{\stopgrad}{\cancel{\nabla}}

\usepackage{dsfont}

\ifarxiv  \fi

\newcommand{\render}{\text{Render}}

\newcommand{\R}[1]{{%
    \textbf{%
        \ifstrequal{#1}{1}{\textcolor{red}{R#1}}{%
        \ifstrequal{#1}{2}{\textcolor{blue}{R#1}}{%
        \ifstrequal{#1}{3}{\textcolor{magenta}{R#1}}{%
        \ifstrequal{#1}{4}{\textcolor{teal}{R#1}}{%
                           \textcolor{cyan}{R#1}%
        }}}}%
    }%
}}

\definecolor{keywordcolor}{rgb}{0.5,0,0.5}
\definecolor{stringcolor}{rgb}{0.7,0,0}
\definecolor{commentcolor}{rgb}{0.25,0.5,0.35}

\definecolor{best}{RGB}{235, 250, 220}

\lstdefinelanguage{jsonlike}{
  morestring=[b]',
  morestring=[b]",
  morecomment=[l]{//},
  literate=
   *{0}{{{\color{blue}0}}}{1}
    {1}{{{\color{blue}1}}}{1}
    {2}{{{\color{blue}2}}}{1}
    {3}{{{\color{blue}3}}}{1}
    {4}{{{\color{blue}4}}}{1}
    {5}{{{\color{blue}5}}}{1}
    {6}{{{\color{blue}6}}}{1}
    {7}{{{\color{blue}7}}}{1}
    {8}{{{\color{blue}8}}}{1}
    {9}{{{\color{blue}9}}}{1}
}

\newcolumntype{Y}{>{\centering\arraybackslash}X}
\newcolumntype{L}[1]{>{\centering\arraybackslash}m{#1}} 
\newcommand{\datasetcol}{12mm} 
\newcommand{\rotlabel}[2][\datasetcol]{%
  \rotatebox[origin=c]{90}{\parbox{#1}{\centering #2}}%
}
\newcommand{\metric}{\tiny PSNR$\uparrow$/SSIM$\uparrow$/LPIPS$\downarrow$}

\makeatletter

\newcounter{supsection}
\renewcommand{\thesupsection}{\ifarxiv A\else S\fi\arabic{supsection}}
\newcounter{supsubsection}[supsection]
\renewcommand{\thesupsubsection}{\ifarxiv A\else S\fi\arabic{supsection}.\arabic{supsubsection}}

\newcommand{\supsection}[1]{%
  \refstepcounter{supsection}%
  \section*{~\thesupsection \quad #1}%
  \addcontentsline{suppcontents}{section}{\protect\numberline{\thesupsection}#1}%
}

\newcommand{\supsubsection}[1]{%
  \refstepcounter{supsubsection}%
  \subsection*{\thesupsubsection\quad #1}%
  \addcontentsline{suppcontents}{subsection}{\protect\numberline{\thesupsubsection}#1}%
}

\newcommand{\listofsuppcontentsname}{Table of Contents}
\newcommand{\listofsuppcontents}{%
  \section*{\listofsuppcontentsname}%
  \@starttoc{suppcontents}%
}

\newcounter{suppfigure}

\newcounter{supptable}

\newenvironment{suppfigure}[1][]{%
  \refstepcounter{suppfigure}%

  \begin{figure}[#1]
  \centering
}{%
  \end{figure}
}

\makeatother

\newcommand{\suptitle}[1]{%
  \begin{center}
    {\Large\bfseries \ifarxiv Appendix \else Supplementary Material \fi} \\[1.5ex]
    {\large #1}\\[1ex]
  \end{center}
}

\makeatletter
\renewcommand{\paragraph}{%
  \@startsection{paragraph}{4}%
  {\z@}{2.0ex \@plus 1ex \@minus .2ex}{-1em}%
  {\normalfont\normalsize\bfseries}%
}
\makeatother  

\usepackage{xr}

\makeatletter
\newcommand*{\addFileDependency}[1]{
  \typeout{(#1)}
  \@addtofilelist{#1}
  \IfFileExists{#1}{}{\typeout{No file #1.}}
}
\makeatother

\newcommand*{\myexternaldocument}[1]{
    \externaldocument{#1}
    \addFileDependency{#1.tex}
    \addFileDependency{#1.aux}
}

\definecolor{cvprblue}{rgb}{0.21,0.49,0.74}
\usepackage[pagebackref,breaklinks,colorlinks,allcolors=cvprblue]{hyperref}
\usepackage[capitalize]{cleveref}
\crefname{section}{Sec.}{Secs.}
\crefname{table}{Table}{Tables}
\crefname{figure}{Fig.}{Figs.}

\ifarxiv \crefname{appendix}{App.}{Apps.}
\else \crefname{appendix}{Suppl.}{Suppls.} \fi

\myexternaldocument{_supplementary}

\begin{document}
\title{\paperTitle}
\author{\authorBlock}



\maketitle

\begin{abstract}
We introduce a probabilistic splat-based radiance field framework that retains the fast rasterization and test-time efficiency of 3D Gaussian Splatting (3DGS) while replacing heuristic primitive manipulation with gradient-based optimization of a volumetric probability density. Rather than relocating, splitting, or culling Gaussians via hand-tuned densification (e.g., ADC), we treat primitive locations as samples drawn from a persistent, learnable density. We instantiate this density using a novel, memory-efficient multi-scale hierarchical grid that enables end-to-end gradient-based optimization. To stabilize the optimization, we derive an unbiased gradient estimator with control variates that markedly reduces variance. By allowing probability mass to flow to where the loss demands, our framework eliminates brittle priors and naturally explores the volume, achieving state-of-the-art reconstruction quality on mip-NeRF 360 while preserving 3DGS-level rendering speed.
\end{abstract}

\section{Introduction}
\label{sec:intro}

Neural radiance fields (NeRFs)~\cite{mildenhall2020nerf} achieve high quality novel view synthesis by optimizing a continuous volumetric density using gradient descent. 3D Gaussian splatting (3DGS)~\cite{3DGS} accelerates rendering by representing scenes with discrete primitives, but it relies on sometimes-brittle heuristics for adding and removing primitives during optimization.

We introduce Eulerian Gaussian splatting (EGS), an Eulerian formulation of splat-based view synthesis that bridges these two approaches, combining the stability of continuous-field optimization with the runtime efficiency of discrete Gaussian primitives. We do this by defining a volumetric probability density function over the scene, and optimizing this field with gradient descent by rendering a discrete set of Gaussians sampled from the field at each iteration.

In contrast to 3DGS and variants such as 3DGS-MCMC~\cite{MCMC}, which take a Lagrangian view by explicitly relocating and restructuring primitives, we adopt an Eulerian perspective and optimize an underlying probability function that governs where Gaussians are allocated. This allows new spatial structures to emerge wherever the loss demands, exploring the volume purely through gradient descent, without any hand-crafted procedures for moving, splitting, or pruning primitives. It bridges the gap between NeRF and 3DGS by producing an efficient 3DGS-style representation at render time, while retaining NeRF-like flexibility to adaptively allocate probability mass during optimization.

\begin{figure}
    \centering
\includegraphics[width=\linewidth]{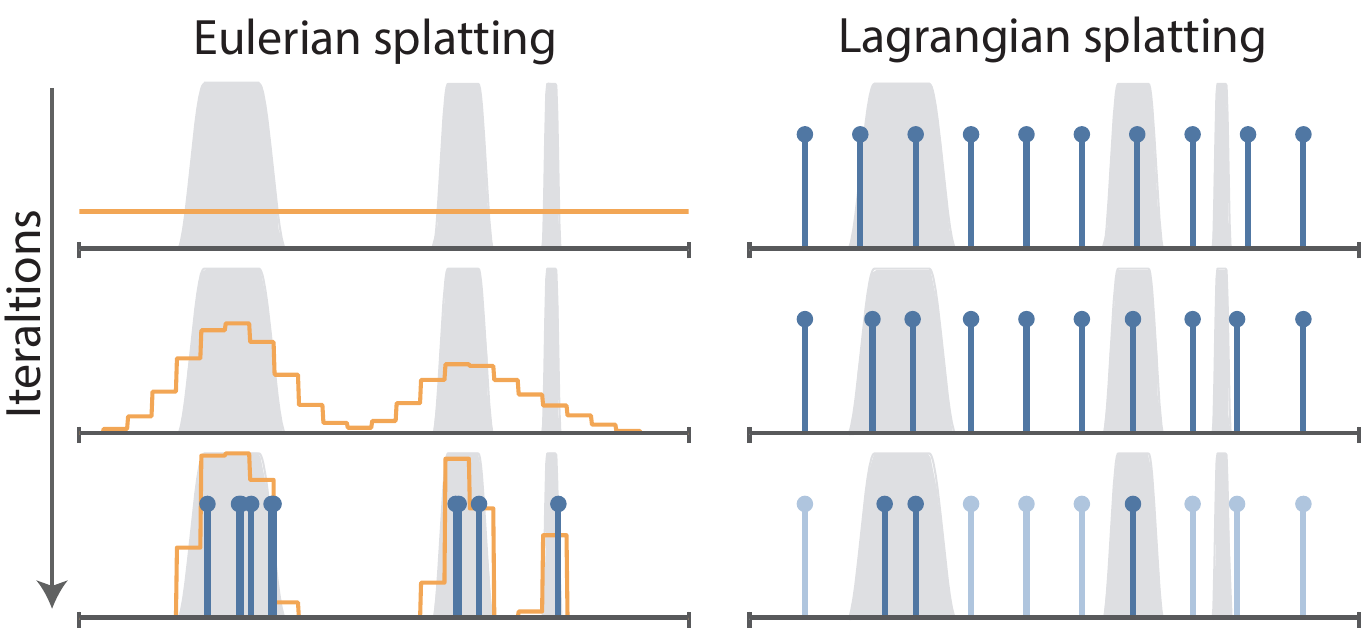}
    \caption{\emph{Left}: We fit to scene shape (gray) by optimizing a probability distribution (orange) that governs primitive locations, creating and removing mass wherever needed, and producing primitives (blue stems) via sampling. \emph{Right}: Previous methods directly optimize primitives, and primitives with no local gradient cannot adapt without additional heuristics for erasure and re-insertion.}
    \label{fig:teaser}
    \vspace{-10px}
\end{figure}

To realize our formulation, we address several challenges. We introduce the \textit{hashed probability pyramid}, a novel volumetric probability representation that achieves high resolution while being memory-efficient, globally normalized, and amenable to efficient sampling. The key idea is to represent the probability density using a multi-scale hierarchical grid, and to exploit scene sparsity by sharing parameters at finer levels of the hierarchy. This allows for resolving thin structures while also being efficient enough for end-to-end optimization on a single GPU.

Optimizing our probability pyramid with stochastic gradient descent introduces another challenge: constructing a gradient estimator with sufficiently low variance to ensure stable learning. We address this by using control variates to derive an unbiased estimator that isolates each Gaussian’s contribution to the rendered images, yielding significantly more informative and less noisy gradient updates.

We experimentally show that our method outperforms all baselines when trained with the same number of primitives initialized from a uniform distribution over the scene volume, and it achieves results comparable to state-of-the-art methods that depend on COLMAP-based initialization.

\section{Related Work}
\label{sec:related}

\paragraph{Radiance fields and hash-encoding grids.}

Radiance fields have become a dominant approach for high-quality novel view synthesis from posed images. Neural radiance fields (NeRFs)~\cite{mildenhall2020nerf, mipnerf360} model a scene as a continuous volumetric function parameterized by an MLP and optimized end-to-end via differentiable volumetric rendering. This has the advantage of allowing optimization by pure gradient decent. However, rendering novel views is slow and computationally-intensive because it requires evaluating MLPs at millions of sample points per image. 

In response, grid-based radiance fields have gained favor because they replace expensive per-sample MLP evaluations with cache-friendly look-ups of precomputed features, enabling orders-of-magnitude speedups in training and rendering~\cite{sun2022direct, karnewar2022relu}. The main drawback of this is memory, which scales quickly with the number of voxels in which features are stored, making high-fidelity reconstructions intractable.

To alleviate memory constraints, sparse hierarchical structures~\cite{yu2021plenoxels, yu2021plenoctrees} and hash encoding grids~\cite{muller2022instant, zipnerf} encode high dimensional feature grids in a compact and scalable manner by compressing dense grids using shared feature tables that are indexed by a hash function. We use existing hash grids to store the attributes of Gaussian primitives (color, scale, \etc), and we create a new hashed hierarchical representation for volumetric probability density that is globally normalized.

\paragraph{Adaptive density control for Gaussian splatting.}
    Another way to perform view synthesis is using discrete primitives. This trades some representational flexibility for explicit control and real-time, hardware-friendly rasterization. 3D Gaussian Splatting (3DGS) represents a scene with a compact set of anisotropic Gaussian primitives and uses differentiable rasterization for real-time rendering~\cite{3DGS}. Its training typically relies on heuristics called ``Adaptive Density Control'' (ADC) to manipulate primitives, for example using the magnitude of positional gradients to identify regions that need more primitives. Similar approaches are used for splatting by other methods based on Lagrangian scene representations~\cite{Held2025Triangle,mai2025ever,govindarajan2025radiant}.

There are efforts to make density control more principled. Taming-3DGS~\cite{taming-3dgs} replaces hand-tuning with a score-based procedure that ranks Gaussians using per-pixel saliency, positional gradients, and primitive attributes. Revising Densification~\cite{revisingdensification} introduces per-pixel error signals and opacity-aware cloning. Efficient Density Control~\cite{deng2025efficient} proposes adaptive rules for splitting and pruning. However, these methods still rely on schedules and thresholds that can be brittle.

\paragraph{Probabilistic approaches to novel view synthesis.}

Challenges with rule-based density control motivates a shift to probabilistic ways of guiding primitives during training. For example, MCMC-3DGS~\cite{MCMC} encourages high-transparency primitives to randomly explore space using Brownian motion and random re-spawn mechanics, while being guided by the probability distribution that is implied by the positions of opaque primitives. This approach has been extended to non-Gaussian primitives as well~\cite{liu2025deformablebetasplatting}. Our method is different because it learns a probability distribution explicitly, eliminating the need for density control altogether. 
 
 
Implicit Neural Point Clouds (INPC)~\cite{INPC} introduces an explicit probability distribution for placing point primitives in a scene. However, their probability field is based on octrees, so its spatial refinement still relies on heuristic subdivision and pruning rules, which can limit flexibility and introduce suboptimal structural decisions.

Another adjacent method is Lagrangian Hashing~\cite{govindarajan2024laghashes}, which compresses neural fields by replacing high-resolution hash encoding grids with a point-based ``Langrangian" representation stored in the upper levels of a hash table. Conceptually, this bridges Eulerian grids and point-based primitives, but unlike our approach it does not learn a normalized spatial probability distribution over primitive locations; instead, capacity is redistributed by an error-driven objective. 

\section{Method}
\label{sec:method}

\begin{figure*}[t]
    \centering
    \includegraphics[width=\textwidth]{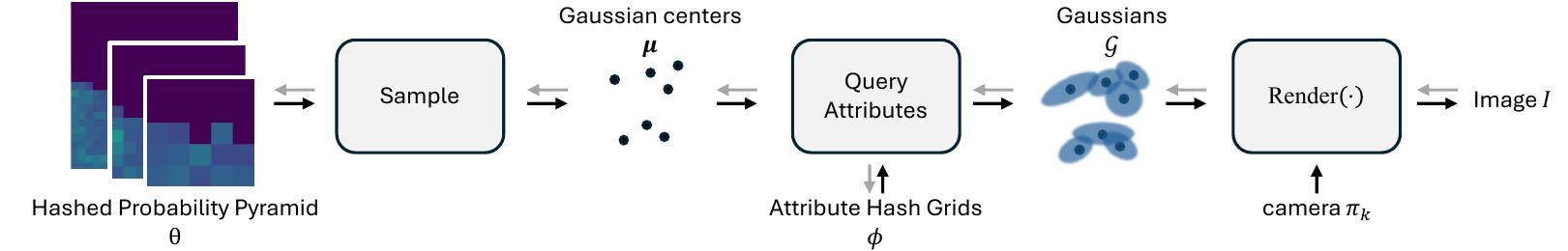}
    \vspace*{-0.3in}
    \caption{We represent the scene as a hierarchical probability density function $p_\theta(\mu)$, along with hash grids $\phi$ that store Gaussian attributes (color, opacity, scale, rotation) at each position. 
    At each training iteration, we sample Gaussian centers $\mu_i\sim p_\theta(\mu)$, look up their attributes $\phi(\mu_i)$, and render the resulting Gaussians with a training camera $\pi_k$ to create image $I$ (black arrows). 
    The gradient of the loss between rendered image $I$ and the corresponding training image is used to update parameters $\theta$ and $\phi$ (gray arrows).}
    \label{fig:pipeline}
\end{figure*}

The input to our method is a dataset of images of a scene with their corresponding known cameras, and our goal is to optimize the scene to match the input images. Our rendering model is probabilistic:
\begin{equation}
\begin{aligned} \label{eq:probmodel}
I = \mathbb{E}_{\mathcal{G}\sim p(\mathcal{G})}\left[\render (\mathcal{G}, \pi)\right],
\end{aligned}
\end{equation}
where $\mathcal{G} = \{\mathcal{G}_i\}_{i=0}^{N-1}$ is a set of $N$ Gaussians sampled from distribution $p(\mathcal{G})$, and $\render(\cdot)$ is the standard 3DGS rasterization which renders the Gaussians using camera $\pi$.

This probabilistic rendering model is the crux of our approach: optimizing $p$ allows our model to efficiently create and remove Gaussians throughout the scene using only gradient cues, while using efficient rasterization for optimization.

In \cref{sec:prob_model} we describe our probabilistic model for $p$, and in \cref{sec:hashed_prob_pyramid} we introduce our novel parameterization and sampling procedure. In \cref{sec:optimization}, we describe how to optimize this new probability data structure.

\subsection{Probabilistic Model} \label{sec:prob_model}

Typical scenes reconstructed by Gaussian splatting require millions of Gaussians, each of which comprises tens of scalars for the Gaussian's position, scale, rotation, opacity, and (view-dependent) color. Modeling and sampling the full distribution $p$ in \cref{eq:probmodel} is prohibitively expensive, so we make some simplifying assumptions in modeling it.

First, we assume that the Gaussians are independent, so
\begin{equation}
    p(\mathcal{G}) = \prod_{i=0}^{N-1} p(\mathcal{G}_i),
\end{equation}
where $p(\mathcal{G}_i)$ is the probability distribution function of a single Gaussian. Second, we split the probability distribution of each Gaussian into a distribution with parameters $\theta$ over the position of the Gaussian, and a distribution with parameters $\phi$ over the remaining Gaussian's attributes:
\begin{equation}
    p(\mathcal{G}_i) = p_\theta(\mu_i)p_\phi(\phi_i|\mu_i),
\end{equation}
where $\mu_i$ is the mean of the $i$th Gaussian, and $\phi_i$ are its remaining attributes (rotation, scale, opacity, and color).

Since the parameters of the attributes $\phi$ are significantly easier to optimize compared with the Gaussian means, we use a deterministic (optimizable) model for $p_\phi(\phi_i|\mu_i)$, realized as a 3D hash grid~\cite{ingp}. The parameters of the hash grid $\phi$ are used to map a Gaussian position $\mu_i$ to its attributes $\phi_i$. 

\subsection{Hashed Probability Pyramid}\label{sec:hashed_prob_pyramid}

\begin{figure}[t]
    \centering
\includegraphics[width=\linewidth]{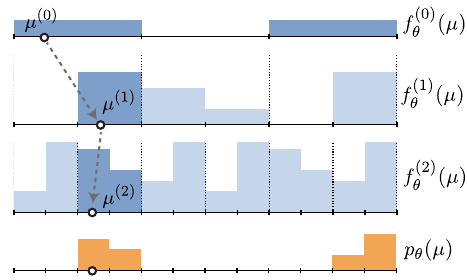}
    \caption{Sampling from a hashed probability pyramid with $L=3$ levels and budget $B=3$. Levels 1 and 2 are normalized within each 2-bin block. Level 2 is hashed: Blocks 1--3 are the same as blocks 4--6. Thus, the pyramid mimics a 12-bin distribution $p_\theta(\mu)$ using only eight degrees of freedom.} 
\label{fig:hash_sampling}
\vspace{-10px}
\end{figure}

We need a probability distribution $p_\theta(\mu_i)$ over 3D space, say $[0, 1]^3$, that has two key properties:  (i) it can be sampled efficiently, and (ii) it supports high resolutions without an infeasible growth in parameters. We achieve this by introducing a \emph{hashed probability pyramid}, which is a memory-efficient, multiscale piecewise-constant distribution function in 3D. Piecewise-constant distributions have been widely used in computer graphics for sampling in one or two dimensions~\cite{pbrt}, and our model extends their benefits to 3D without a cubic growth in memory.

We parameterize the full distribution as a product of functions over $L$ levels:
\begin{align} \label{eq:ppyramid}
    p_\theta(\mu) = \frac{1}{Z} \prod_{\ell=0}^{L-1} f^{(\ell)}_\theta(\mu),
\end{align}
where $Z$ is a normalizing factor. Each $f^{(\ell)}_\theta$ is a piecewise-constant function with $N_\ell\times N_\ell\times N_\ell$ bins that partition the volume:
\begin{equation}
    f^{(\ell)}_\theta(\mu) = \sum_{\volindex\in \mathcal{I}_\ell} \theta^{(\ell)}_{\volindex}\mathds{1}\left[\lfloor\mu N_{\ell}\rfloor=\volindex\right],
\end{equation}
where $\mathcal{I}_\ell = \{0, ..., N_\ell-1\}^3$ is the set of 3D indices of the level-$\ell$ grid, and $\theta^{(\ell)}_{\volindex}$ is a parameter representing the unnormalized probability density associated with index $\volindex$ in level $\ell$. The expression $\mathds{1}\left[\lfloor\mu N_{\ell}\rfloor=\volindex\right]$ partitions the volume $[0, 1]^3$ into bins; it returns $1$ if $\mu$ is in the bin with index $\volindex$ and $0$ otherwise. In our pyramid, level $\ell=0$ is coarsest (lowest resolution) and level $\ell=L-1$ is finest (highest resolution). We assume for simplicity that resolutions grow by a factor of two, \ie, $N_{\ell+1} = 2N_\ell$. 

The $0$th function $f^{(0)}_{\theta}$ is normalized to integrate to $1$, and the other functions $f^{(1)}_\theta, ..., f^{(L-1)}_{\theta}$ are normalized such that each of their $2\times 2\times 2$ blocks integrates to $1$. This makes the parameterization \emph{complete}, meaning that any $p(\mu)$ can be written as a product of such functions without introducing additional degrees of freedom.  We see this by tallying the pyramid's degrees of freedom:
\begin{equation}
    N_0^3-1+\sum_{\ell=1}^{L-1}((N_{\ell+1}/N_{\ell})^3-1)\cdot N_\ell^3=N_{L-1}^3-1,
\end{equation}
which is the same as that of a standard distribution with $N_{L-1}\times N_{L-1}\times N_{L-1}$ bins.

However, naïvely using $N_{L-1}^3-1$ parameters to represent our distribution is infeasible for high resolution grids. To address this, we upper bound the number of parameters in each level. We allocate at most $B$ distinct blocks that are $2\times 2\times 2$ at each level, and we randomly tile them across the level's bins using a hash function $T_\ell\colon \{0, ..., N_\ell-1\}^3\rightarrow\{0,...,B-1\}$. This populates the $N_\ell \times N_\ell\times N_\ell$ bins at the level using only $B$ distributions, with the bin whose index is $\volindex \in \{0, ..., N_{\ell-1}\}^3$ being populated by a value from the $2\times 2\times 2$ block given by $T_\ell(\volindex)$. 
\Cref{fig:hash_sampling} shows an example in one dimension.

Hashing greatly reduces the parameter count without significantly impacting representational capacity. There can be ``hash collisions'' for pyramid levels with $N_\ell > 2B^{1/3}$, meaning that multiple bins from level $\ell-l$ are mapped to the same block at level $\ell$; but the sparsity of surfaces in three dimensions means that these collisions often occur in empty parts of the volume which, as depicted in \cref{fig:hash_sampling}, can easily be accommodated by setting coarser bins to zero. Note that hash collisions are independent for each level, so the probability of the same collision occurring across all levels is vanishingly small. These are similar to the motivations for Instant-NGP~\cite{muller2022instant}, and in fact we use their hash encoding for our $T_\ell$ maps. See the appendix for details.

\subsubsection{Sampling from a Hashed Probability Pyramid}
\label{sec:sampling}

Another requirement of our representation is efficient sampling. This is important for  optimizing the distribution parameters $\theta$ and Gaussian attributes $\phi$ based on the probabilistic rendering model in \cref{eq:probmodel}. Fortunately, sampling from the hashed probability pyramid is trivial: We sample from the coarsest distribution and then iteratively refine the sample by traversing levels conditioned on the previous bin. 

Formally, we generate a sample $\mu^{(L-1)} \sim p_\theta(\mu)$ by:
\begin{align} \label{eq:sampling}
    \mu^{(0)} &\sim p^{(0)}_\theta(\mu),\\
    \mu^{(\ell)} &\sim p^{(\ell)}_\theta\left(\mu \;\Big| \;\lfloor \mu^{(\ell-1)} N_{i-1}\rfloor\;\right), \text{ for } \ell=1,...,L-1, \nonumber
\end{align}
where $p^{(0)}_\theta(\mu) = \frac{1}{Z_0}f^{(0)}_\theta(\mu)$ is the distribution corresponding to the coarsest level of the pyramid in \cref{eq:ppyramid} ($Z_0$ is the normalization factor), and the $\ell$th conditional distribution $p_\theta^{(\ell)}(\mu | \cdot)$ for $\ell=1,...,L-1$ is the $2\times 2\times 2$ piecewise-constant distribution that subdivides the bin that was sampled at level $\ell-1$. See \cref{fig:hash_sampling}.

Each of the sampling stages in \cref{eq:sampling} can be done efficiently with standard inverse transform sampling, which passes a  three-vector $u^{(\ell)}\sim \mathrm{U}([0, 1]^3)$ through the appropriate inverse cumulative distribution function for each stage. Also, we make the sequence of $L$ sampling stages end-to-end auto-differentiable by tying the stages together via: (i) drawing a single random sample $u^{(0)}\sim \mathrm{U}([0, 1]^3)$; and then (ii)  recursively defining subsequent three-vectors using deterministic maps:
\begin{equation} \label{eq:differentiablesampling}
    u^{(\ell)} = \operatorname{frac}\left(\mu^{(\ell-1)}N_{\ell-1}\right) \text{ for } \ell= 1,..., L-1,
\end{equation}
where $\operatorname{frac}(x)=x-\lfloor x\rfloor$ is the fractional part function. This works because all of our distributions are piecewise-constant, implying that the distribution within each bin is uniform, and thus that \cref{eq:differentiablesampling} yields valid samples from $\mathrm{U}([0, 1]^3)$ for all levels $\ell$. This approach guarantees the differentiability of the final sample position $\mu^{(L-1)}$ with respect to all of the pyramid parameters that led to it. 

\subsubsection{Handling Unbounded Scenes}\label{subsec:unbounded}

Our hashed probability pyramid provides samples in $[0, 1]^3$, but 3D scenes are often unbounded. In order to map the samples to all of 3D space $\mathbb{R}^3$, we first apply an affine mapping to map our samples to $[-1, 1]^3$ and then we use a contraction function $\mathcal{C}:[-1, 1]^3\rightarrow\mathbb{R}^3$ defined as:
\begin{equation}\label{eq:xinvcontract}
    \mathcal{C}(\mu)= \begin{cases}
\frac{\mu}{a}, & \|\mu\|_\infty \le a, \\
\frac{1-a}{1-\|\mu\|_\infty}\frac{\mu}{\|\mu\|_\infty}, & \text{otherwise},
\end{cases}
\end{equation}
where $\|\mu\|_\infty=\max_n|\mu_n|$ denotes the $L_\infty$ norm, and $a$ is a contraction factor that controls how the model's capacity is allocated to near and far content. (We set $a=\nicefrac{3}{4}$ in our experiments.)

Our contraction function is a modification of the (inverse) contraction function from mip-NeRF~360~\cite{mipnerf360}, but modified to have $L_\infty$ instead of $L_2$ symmetry, since our distribution is defined using a cubic grid.

\section{Optimization}\label{sec:optimization}

Our optimization task is to adjust distribution parameters $\theta$ and Gaussian attributes $\phi$ to minimize the error between a set of observed images $\{\IGT_k\}_{k=0}^{K-1}$ and the corresponding renderings $\{I_k\}_{k=0}^{K-1}$ as described in \cref{sec:method}. The loss can be written as

\begin{equation}
    \sum_{k=0}^{K-1} \mathcal{L}(I_k, \IGT_k),
\end{equation}
where $\mathcal{L}$ is the loss between one pair of images.

Minimizing the loss by gradient descent requires estimating the gradient with respect to $\xi \in \{\theta, \phi\}$. We write:
\begin{align}
\label{eq:goal}
&\nabla_\xi \mathcal{L}(I, \IGT)
=\nabla_{I} \mathcal{L}(I, \IGT) \cdot \nabla_\xi I,
\\
&\text{with } I = \mathbb{E}_{\bmu \sim p_\theta(\bmu)}\left[\render (\bmu, \phi(\mu), \pi_k)\right].\nonumber
\end{align}
Here, $\render(\cdot)$ is the 3DGS rasterization~\cite{3DGS}, from camera $\pi_k$, of the set of $M$ Gaussians with means $\{\mu_i\}_{i=0}^{M-1}$ that are sampled according to \cref{sec:hashed_prob_pyramid}. Within the rendering operator, the Gaussian attributes $\phi(\mu_i)$ are obtained by querying attribute hash grids~\cite{ingp} at each location $\mu_i$.

In all of our experiments, the expected values for an image $I$ and image gradients $\nabla_\xi I$ in \cref{eq:goal} are estimated using Monte Carlo estimation with a single sample from the joint distribution $p_\theta(\mu)$, \ie, at every iteration we use a single set of $N$ Gaussians sampled from the hashed probability pyramid. As is usual for optimization using Monte Carlo sampling, our goal is to find an unbiased estimator for the gradients and to reduce its variance to make optimization faster and more robust.

\subsection{Gradient Estimators and Variance Reduction}\label{sec:estimator}

The gradient in \cref{eq:goal} contains two terms. The first term is the gradient of the loss function with respect to the image, which is trivial to compute. The second term is also trivial to compute using automatic differentiation for the Gaussian attributes $\phi$. However, since $\theta$ affects the distribution used for computing the expectation, its gradient $\nabla_\theta I$ requires more care, and the variance of different unbiased estimators can significantly change the optimization behavior. 

Our sampling procedure from \cref{sec:hashed_prob_pyramid} is  differentiable, so an obvious way to estimate $\nabla_\theta I$ would be to use automatic differentiation, equivalent to the well-known \emph{pathwise estimator}. This is unbiased, but as we show in the appendix and through ablations in \cref{sec:ablations}, it has high variance in practice, leading to slow training and poor convergence. 

Instead, we design a new unbiased gradient estimator for $\nabla_\theta I$. It is based on the standard score function estimator,
\begin{equation}
\nabla_\theta I
=\mathbb{E}_{\mu\sim p_\theta(\mu)}\left[I\cdot  \sum_{i=0}^{M-1}\nabla_\theta\log p_\theta(\mu_i)\right],
\label{eq:scoreestimator}
\end{equation}
where the summation is over all Gaussians $i=0,...,M-1$. We use control variates to modify it to:
\begin{equation}
\nabla_\theta I
=\mathbb{E}_{\mu\sim p_\theta(\mu)}\left[
\sum_{i=0}^{M-1}(I-I_{-i})\cdot\nabla_\theta\log p_\theta(\mu_i)\right],
\label{eq:leave-one-out}
\end{equation}
where $I_{-i}$ is the image rendered with $M-1$ Gaussians, excluding the $i$th one. This greatly reduces variance by weighting the gradient contribution from each sample $\mu_i$ by its individual impact on the image. Also, the estimator remains unbiased because $I_{-i}$ and $\mu_i$ are statistically independent, and because $\mathbb{E}[\nabla_\theta \log p_\theta]=0$ and therefore $\mathbb{E}[I_{-i}\cdot\nabla_\theta \log p_\theta(\mu_i)]=0$. See the appendix for additional details and intuition.

At first glance, the estimator in \cref{eq:leave-one-out} seems impossible to use in practice, because it suggests rendering $M$ images, one for each omitted Gaussian. But surprisingly, the differences $(I-I_{-i})$ are readily computed during auto-differentiation for the gradients with respect to the Gaussian attributes $\phi$. Indeed, in the appendix we prove that:
\begin{equation} \label{eq:diffimg}
I - I_{-i} = o_i \frac{\partial I}{\partial o_i},
\end{equation}
where $o_i$ is the opacity of the $i$th Gaussian. This means that our unbiased estimator can be computed using the right side of \cref{{eq:diffimg}}, providing similar computational efficiency as the gradients that would be obtained by automatic differentiation, but with much lower variance.

\subsection{Loss Function}
\label{sec:loss}

We follow 3DGS and use a combination of $L_1$ and D-SSIM losses. In addition, inspired by 3DGS-MCMC~\cite{MCMC}, we apply $L_1$ regularization to the opacities $o_i$ and scales $s_i$. To encourage the model to prioritize lower order spherical harmonic coefficients, we apply a loss to the spherical harmonics coefficients $c_{i,\ell}$ of order $\ell \geq 1$. In total, the loss for the $k$th image is:
\begin{align}
&\mathcal{L} = \lambda_1\|I-\IGT||_1 + \,(1-\lambda_1) \text{SSIM}(I,\IGT) \\
&+\sum_{i \in \mathcal{X}_k} \left(\lambda_o |o_i|\mathds{1}[o_i>\tau] \, + \,\lambda_{s} \|s_{i}\|_1 \, + \, \lambda_c  \|w\odot c_{i}\|_1\right), \nonumber
\end{align}
where $\mathcal{X}_k$ denotes the Gaussians within the view frustum of camera $\pi_k$, $\tau=0.05$ is a threshold for the opacity loss, and $w$ is a weighting term for the different spherical harmonics terms, which we set to $0$ for $\ell=0$ and $0.2^\ell$ for $\ell \geq 1$. We set $\lambda_1 =0.8$, $\lambda_o=0.05$, $\lambda_s=0.02$, and $\lambda_c = 10^{-3}$ for all of our experiments.

\section{Implementation details}

\subsection{Optimizing the distribution} \label{sec:dist_opt}

Due to the stochastic representation of the scene's geometry, we need to take extra care when sampling from the distribution and when optimizing its parameters $\theta$.

\paragraph{Defensive Sampling.}

A well-known problem with gradient-based optimization of distributions using sampling is that if at any point a certain part of the distribution incorrectly reaches zero probability density, the model will never be able to recover. To avoid this we add noise to our sampling process, specifically adding Gaussian noise to 20\% of the samples prior to mapping them to world space using~\cref{eq:xinvcontract}. Our Gaussian noise has a standard deviation of $2\cdot10^{-3}$ at the beginning of training, and it is linearly annealed to zero over the first twenty thousand iterations.

We find that our model already does a good job of localizing mass without defensive sampling due to the hierarchical nature of the probability grids, but the added noise early in training helps our model explore space over more iterations.

\paragraph{Sample Rounding.}

Randomness in the exact positions of the samples could cause the distances of the Gaussians from the camera to change, which causes slight changes in the renderings and introduces variance in our estimators. In order to resolve this and lower variance, we ``round'' the position of every sample to lie at the center of its bin at the highest grid resolution:
\begin{equation} \label{eq:rounding}
    \mu' = \frac{\lfloor\mu N_{L-1}\rfloor + \nicefrac{1}{2}}{N_{L-1}}.
\end{equation}
Note that this does not affect our gradients since we use our gradient estimator from \cref{sec:estimator}.

\paragraph{Duplicate Sample Removal.}

Another source of variance in the image and gradient estimators arises from the nonlinearity of the alpha compositing operation used for rendering. Because of our sample rounding, multiple samples in the same bin will overlap perfectly with one another, affecting their overall opacity. We reduce the variance caused by this overlap by removing duplicate samples: After rounding the samples using~\cref{eq:rounding}, we discard duplicate positions and only render the unique set of Gaussian centers.

This process has another advantage. As the model optimizes, it tends to place larger probabilities at parts of the scene that need it. This causes the number of duplicate samples to increase, which means that after duplicate removal, the number of Gaussian we actually render decreases over time. This gives our model a natural way of pruning the number of Gaussians during optimization. In our experiments we start optimization with $N=1.5\cdot 10^{7}$ Gaussians, which the model lowers to an automatically-determined number. In order to make our comparisons fair, we set a lower bound on the number of Gaussians that is equal to the number of primitives reported by 3DGS-MCMC~\cite{MCMC}. We do this by resampling from $p_\theta$ when the number of unique primitives in our collection of sampled Gaussians drops below this threshold.

\paragraph{Final Refinement.}

After optimizing the probability and attribute grids, we sample the distribution and evaluate the attributes to get a final collection of Gaussians, each with its own mean and set of attributes. We then refine these Gaussians by running 5000 iterations of standard gradient-based optimization of the primitive locations and attributes. This slightly improves the rendering quality of our method. We report the results with and without this refinement step in our experiments.

\subsection{Gaussian Attribute Grid Settings}\label{sec:representing_attribs}

The Gaussian attributes $\phi$ are represented as a multi-resolution hash encoded grid.
The grids have 17 feature channels total, which are split into three parts, each of which is mapped by a small MLP to features representing opacity (using $1$ channel), color ($8$ channels), scale and rotation ($8$ channels for both attributes). Importantly, because of sample rounding, the value queried inside every bin of the distribution is constant, which lowers the variance of our estimators. See the appendix for additional details.

We find that when optimizing  Gaussian scales, distant Gaussians tend to be small, leading to wasted capacity and a need for many Gaussians to represent the background. To resolve this, we multiply the scale parameters by the inverse square root of the Jacobian of the contraction function, as done in NeRF-Casting~\cite{verbin2024nerf}. This factor ensures that Gaussians whose optimizable scale parameters are the same will project to approximately the same-sized 2D Gaussians, regardless of their distance from the center of the scene. 

\subsection{Hashed Probability Pyramid Settings}
\label{sec:probpyramidsettings}

Our hashed probability pyramid has $L=12$ levels. The coarsest one, $p^{(0)}_\theta$, has resolution $N_0=2$, and the finest one has resolution $N_{11}=N_0\cdot 2^{11}=4096$. We set a budget of $B=2^{18}$ distributions of size $2\times 2\times 2$ , meaning that each level has at most $2^{21}$ parameters. This means that the first levels $p^{(0)}_\theta, ..., p^{(6)}_\theta$ are dense (\ie, they have no parameter sharing), and levels $p^{(7)}_\theta, ..., p^{(11)}_\theta$ contain $B$ distributions of size $2\times 2\times 2$ which tile their respective grids.

\subsection{Scene Scale Normalization}

Similar to mip-NeRF~360~\cite{mipnerf360}, we normalize the camera centers such that their principal components align with the world $X$, $Y$, and $Z$ axes, and the camera centers are contained within the cube $[-1,1]^3$. This normalization transformation is estimated using the training cameras and then applied to the testing cameras during evaluation to ensure alignment. 

\subsection{Additional training details}

To avoid unnecessary computations and speed up training, we cull Gaussians whose centers lie outside the current camera view frustum (within a small margin). This reduces hash-grid queries to only Gaussians that can contribute to the rendered image. 

We train our model with uniformly-sampled random background colors $c_\text{bg}\in[0, \half]^3$ and evaluate the model with black background $c_\text{bg}=0$. We observe that these darker, randomized backgrounds help in early training because the model can use the background to approximate shadows, which temporarily frees up capacity to allocate elsewhere without pushing opacity or probability to zero for Gaussians representing black image pixels.

\section{Experiments}
\label{sec:experiments}

\subsection{Datasets and Metrics}

We evaluate our method on all scenes from mip-NeRF~360~\cite{mipnerf360}, as well as two scenes from Tanks \& Temples~\cite{TANT} and two scenes from Deep Blending~\cite{DeepBlending2018}. We report standard image-based metrics for novel view synthesis: Peak Signal-to-Noise Ratio (PSNR), Structural Similarity Index (SSIM)~\cite{wang2004SSIM}, and Learned Perceptual Image Patch Similarity (LPIPS)~\cite{zhang2018lpips}. 

\subsection{Training}

Empirically, we observe that outdoor mip-NeRF~360 scenes converge faster than indoor or other dataset scenes. Therefore we train outdoor scenes for 35K total iterations (30K iterations of our probabilistic model and 5K refinement iterations, see \cref{sec:dist_opt}), and indoor scenes for 65K iterations (60K probabilistic and 5K refinement steps). All experiments are conducted on a single NVIDIA H200 GPU, requiring approximately 3.5 hours for 35K iterations and 7.25 hours for 65K iterations. 

\subsection{Baselines}

We report results for 3DGS-MCMC as a state-of-the-art baseline, and for Taming-3DGS~\cite{taming-3dgs} as a proxy for standard 3DGS, since it allows training with a fixed budget of Gaussians. We fix the total number of training iterations across methods, training outdoor mip-NeRF~360 scenes for 35K iterations and all other scenes for 65K iterations. If a baseline trained for 65K performs better at 35K than 65K, we report those values instead. Models under the column label COLMAP in~\cref{tab:results} initialize Gaussian centers at the 3D points produced by COLMAP's SfM model. In contrast, models with the column label Random, like ours, initialize Gaussians by sampling uniformly over the volume. For random initialization, we adapt Taming-3DGS to follow the 3DGS-MCMC procedure of sampling 100K points uniformly within a cube bounded by the camera centers. Our primary point of comparison is 3DGS-MCMC with random initialization (MCMC-Random); since it shares our method's initialization strategy and training budget, it enables a direct assessment of modeling and performance differences.

\subsection{Results}

We report the average of each metric for each dataset in \cref{tab:results} along with qualitative examples in \cref{fig:results}. Additional qualitative examples are in the appendix. Our method achieves state-of-the-art performance among randomly initialized models, outperforming all prior approaches in overall dataset PSNR. Compared to models that are initialized using structure from motion, our approach attains competitive results, closing much of the performance gap despite starting from random initialization. 

This demonstrates that our sampling-based training procedure can recover high-quality geometry and appearance purely from image supervision. Overall, these results highlight the effectiveness and robustness of our approach: It does not require SfM points for initialization, yet it achieves rendering quality comparable to methods that require them. 

This enables a simple, heuristic-free, end-to-end training procedure that generalizes across diverse scenes.

\definecolor{best}{RGB}{235, 250, 220}     
\definecolor{second}{RGB}{255, 240, 200}   
\definecolor{highlight}{RGB}{255, 220, 220} 

\begin{table*}[t] 
\centering
\footnotesize
\setlength{\tabcolsep}{5pt}
\renewcommand{\arraystretch}{1.15}

\setlength{\tabcolsep}{2pt} 

\begin{tabularx}{\textwidth}{L{\datasetcol}@{}l@{\hspace{1pt}}Y@{\hspace{1pt}}Y!{\color{black}\vrule width 1.2pt}Y@{\hspace{1pt}}Y@{\hspace{1pt}}Y@{\hspace{1pt}}Y@{\hspace{1pt}}Y}
\cmidrule[1pt](l{0pt}r{0pt}){1-4}\cmidrule[1pt](l{0pt}r{0pt}){5-8}
\multirow{3}{*}{} &
\multirow{3}{*}{} &
\multicolumn{2}{c}{\textbf{COLMAP}} &
\multicolumn{4}{c}{\textbf{Random}} \\
\cmidrule(l{10pt}r{2pt}){3-4}\cmidrule(l{0pt}r{0pt}){5-8}
&& \textbf{Taming-3DGS} & \textbf{MCMC} & \textbf{Taming-3DGS} & \textbf{MCMC} & \textbf{Ours} & \textbf{Ours (refined)} \\
&& \metric & \metric & \metric & \metric & \metric & \metric \\
\cmidrule(l{0pt}r{0pt}){3-4}\cmidrule(l{0pt}r{0pt}){5-8}

\multirow{9}{*}{\rotatebox{90}{\textbf{Mip-NeRF 360~\cite{mipnerf360}}}}
 & Bicycle    & 25.89 / 0.79 / 0.21 & 
                26.18 / 0.81 / 0.18 & 
                24.26 / 0.68 / 0.33 & 
                26.08 / 0.81 / 0.19 & 
                26.12 / 0.80 / 0.23 & 
                26.31 / 0.80 / 0.22 \\ 
 & Garden     & 28.19 / 0.88 / 0.11 & 
                28.23 / 0.88 / 0.10 & 
                26.60 / 0.83 / 0.16 & 
                27.97 / 0.88 / 0.11 & 
                28.02 / 0.88 / 0.13 & 
                28.34 / 0.88 / 0.12 \\  
 & Stump      & 26.88 / 0.78 / 0.23 & 
                27.48 / 0.82 / 0.19 & 
                19.29 / 0.46 / 0.46 & 
                27.39 / 0.81 / 0.20 & 
                27.22 / 0.81 / 0.22 & 
                27.35 / 0.81 / 0.21 \\ 
 & Flowers    & 22.17 / 0.63 / 0.35 & 
                22.29 / 0.66 / 0.27 & 
                20.79 / 0.57 / 0.37 & 
                22.21 / 0.66 / 0.27 & 
                21.97 / 0.65 / 0.32 & 
                22.22 / 0.65 / 0.31 \\ 
 & Treehill   & 23.21 / 0.66 / 0.35 & 
                23.20 / 0.67 / 0.27 & 
                22.18 / 0.62 / 0.40 & 
                23.17 / 0.67 / 0.27 & 
                23.06 / 0.67 / 0.34 & 
                23.23 / 0.67 / 0.33 \\ 
\cmidrule(l{0pt}r{0pt}){3-4}\cmidrule(l{0pt}r{0pt}){5-8}
 & Kitchen    & 32.42 / 0.94 / 0.13 & 
                32.54 / 0.94 / 0.13 & 
                31.23 / 0.93 / 0.15 & 
                32.45 / 0.94 / 0.14 & 
                32.19 / 0.93 / 0.15 & 
                32.48 / 0.93 / 0.15 \\ 
 & Counter    & 29.84 / 0.92 / 0.23 & 
                29.80 / 0.93 / 0.21 & 
                28.86 / 0.89 / 0.28 & 
                29.60 / 0.92 / 0.22 & 
                29.47 / 0.92 / 0.23 & 
                29.65 / 0.92 / 0.23 \\ 
 & Bonsai     & 33.44 / 0.95 / 0.22 & 
                33.12 / 0.96 / 0.21 & 
                31.92 / 0.93 / 0.24 & 
                32.94 / 0.95 / 0.21 & 
                32.75 / 0.95 / 0.23 & 
                33.15 / 0.95 / 0.22 \\ 
 & Room       & 32.70 / 0.93 / 0.25 & 
                32.58 / 0.94 / 0.23 & 
                31.04 / 0.91 / 0.30 & 
                32.50 / 0.94 / 0.24 & 
                32.09 / 0.93 / 0.26 & 
                32.41 / 0.93 / 0.25 \\ 
\cmidrule(l{0pt}r{0pt}){3-4}\cmidrule(l{0pt}r{0pt}){5-8}
 & \textbf{Average} & 28.30 / 0.83 / 0.23 & 
                      28.38 / 0.84 / 0.20 & 
                      26.24 / 0.76 / 0.30 & 
                      28.26 / 0.84 / 0.21 & 
                      28.10 / 0.84 / 0.23 & 
                      28.35 / 0.84 / 0.23 \\ 
 & \textbf{Avg Outdoor} & 25.27 / 0.75 / 0.25 & 
                          25.48 / 0.77 / 0.20 & 
                          22.62 / 0.63 / 0.34 & 
                          25.36 / 0.77 / 0.21 & 
                          25.28 / 0.76 / 0.25 & 
                          25.49 / 0.76 / 0.24 \\ 
& \textbf{Avg Indoor} & 32.1 / 0.93 / 0.21 & 
                        32.01 / 0.94 / 0.20 & 
                        30.76 / 0.91 / 0.24 & 
                        31.87 / 0.94 / 0.20 & 
                        31.62 / 0.93 / 0.22 & 
                        31.92 / 0.93 / 0.21 \\ 
\cmidrule[0.75pt](l{0pt}r{0pt}){2-4}\cmidrule[0.75pt](l{0pt}r{0pt}){5-8}

\multirow{4}{*}{\rotlabel{\textbf{T\&T ~\cite{TANT}}}}
 & Train      & 22.94 / 0.83 / 0.23 & 
                22.94 / 0.84 / 0.21 & 
                21.96 / 0.78 / 0.28 & 
                22.25 / 0.82 / 0.24 & 
                22.50 / 0.81 / 0.24 & 
                22.53 / 0.81 / 0.24 \\ 
 & Truck      & 26.12 / 0.89 / 0.15 & 
                26.38 / 0.90 / 0.13 & 
                23.56 / 0.84 / 0.21 & 
                26.10 / 0.89 / 0.13 & 
                26.20 / 0.88 / 0.17 & 
                26.28 / 0.89 / 0.17 \\ 
\cmidrule(l{0pt}r{0pt}){3-4}\cmidrule(l{0pt}r{0pt}){5-8}
 & \textbf{Average} & 24.53 / 0.86 / 0.19 & 
                      24.66 / 0.87 / 0.17 & 
                      22.76 / 0.81 / 0.25 & 
                      24.17 / 0.85 / 0.19 & 
                      24.35 / 0.84 / 0.20 & 
                      24.40 / 0.85 / 0.20 \\ 
\cmidrule[0.75pt](l{0pt}r{0pt}){2-4}\cmidrule[0.75pt](l{0pt}r{0pt}){5-8}

\multirow{4}{*}{\rotlabel{\textbf{DB ~\cite{DeepBlending2018}}}}
 & Dr Johnson & 29.49 / 0.90 / 0.31 & 
                29.43 / 0.90 / 0.30 & 
                28.84 / 0.89 / 0.34 & 
                29.12 / 0.89 / 0.31 & 
                28.98 / 0.90 / 0.32 & 
                29.21 / 0.90 / 0.31 \\ 
 & Playroom   & 29.76 / 0.90 / 0.30 & 
                29.78 / 0.90 / 0.30 & 
                29.69 / 0.90 / 0.31 & 
                30.32 / 0.91 / 0.29 & 
                30.25 / 0.91 / 0.30 & 
                30.49 / 0.91 / 0.30 \\ 
\cmidrule(l{0pt}r{0pt}){3-4}\cmidrule(l{0pt}r{0pt}){5-8}
 & \textbf{Average} & 29.62 / 0.90 / 0.31 & 
                      29.60 / 0.90 / 0.30 & 
                      29.26 / 0.90 / 0.33 & 
                      29.72 / 0.90 / 0.30 & 
                      29.61 / 0.91 / 0.31 & 
                      29.85 / 0.91 / 0.30 \\ 
\cmidrule[1pt](l{0pt}r{0pt}){1-4}\cmidrule[1pt](l{0pt}r{0pt}){5-8}
\vspace{-10px}
 \end{tabularx}

\caption{Reconstruction metrics of our method compared with Taming-3DGS~\cite{taming-3dgs} and 3DGS-MCMC~\cite{MCMC} on the mip-NeRF 360~\cite{mipnerf360}, Tanks \& Temples (T\&T)~\cite{TANT} and Deep Blending (DB)~\cite{DeepBlending2018} datasets. We show the results using random initialization on the right, and using COLMAP points on the left. We include results for our model before the refinement procedure described in~\cref{sec:dist_opt} (second to last column) as well as after 5K steps of refinement (last column). 
}

\label{tab:results}
\vspace{-10px}
\end{table*}

\begin{figure}
    \centering
\includegraphics[width=\linewidth]{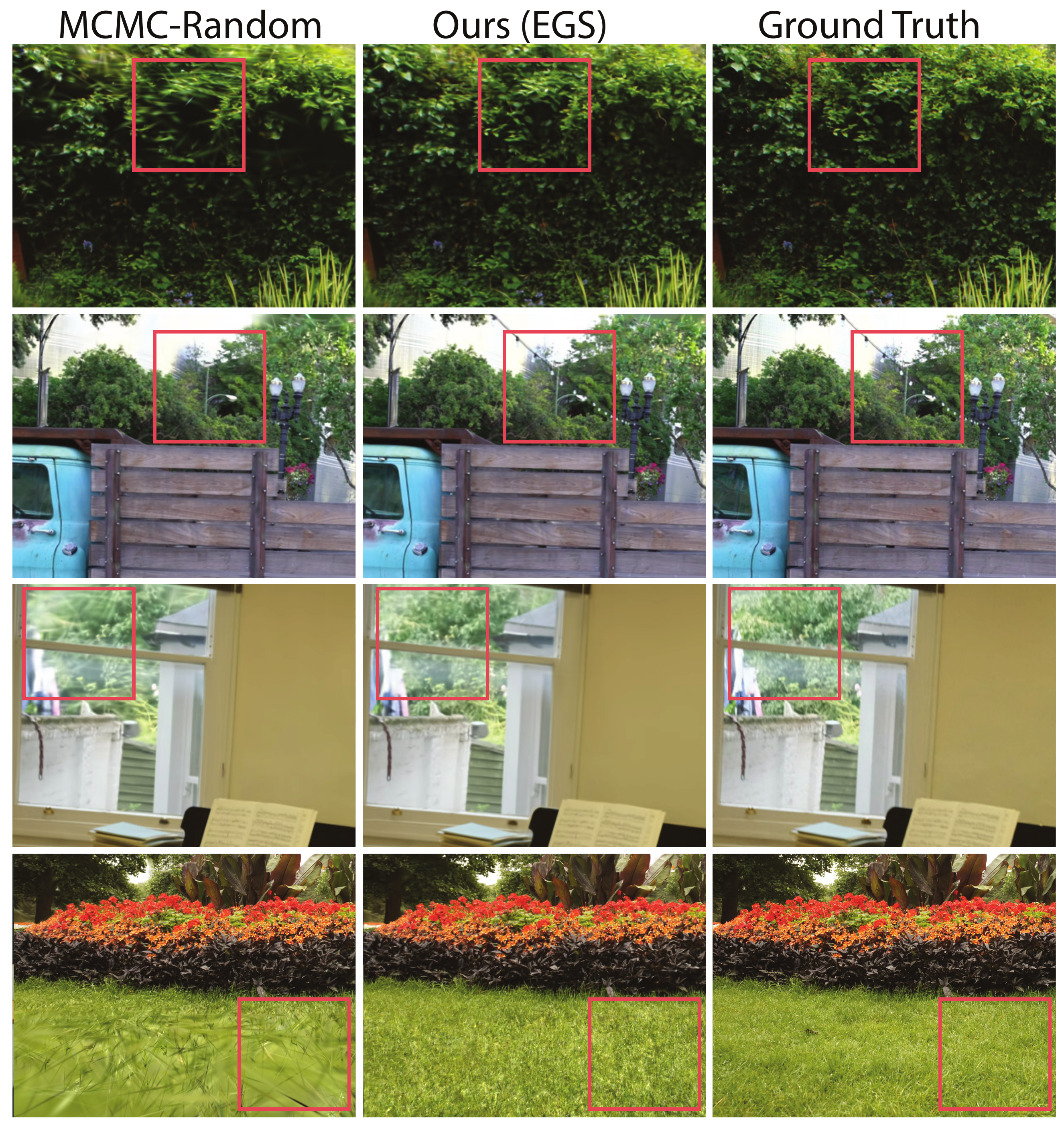}
    \caption{Qualitative comparison between MCMC-Random, our method (EGS), and ground truth. Our model more faithfully captures structural and global scene details. For example, it reconstructs the string lights (row 2) and distant trees (row 3) that MCMC-Random misses. It also captures finer details in the leaves (row 1) and grass (row 4), with fewer visible artifacts.}
    \label{fig:results}
    \vspace{-10px}
\end{figure}

\subsection{Ablations}\label{sec:ablations}

We validate our design choices and quantify their impact on performance in \cref{tab:ablations}. The most influential component is our custom gradient estimator. Without it, training becomes significantly less stable and converges to much worse results.

Opacity regularization also provides a major benefit. Without it, large opaque Gaussians occlude signal beneath them and training stalls. (See qualitative comparisons in the appendix.) Sample rounding provides a clear advantage before refinement, but the advantage narrows after refinement. This is expected because variance reduction helps during training, but once Gaussians are represented as discrete primitives, refinement can optimize small discrepancies directly. Finally, scaling regularization and defensive sampling each offer modest gains to model performance and additionally benefit training stability. 

\begin{table}[h]
\centering
\footnotesize
\setlength{\tabcolsep}{3pt}
\renewcommand{\arraystretch}{1.15}

\begin{tabularx}{\linewidth}{@{}l *{2}{Y} @{}}
\toprule
\multirow{2}{*}{\textbf{Configuration}} &
\multicolumn{1}{c}{\textbf{Pre-refinement}} &
\multicolumn{1}{c}{\textbf{Post-refinement}} \\
& \tiny PSNR$\uparrow$ / SSIM$\uparrow$ / LPIPS$\downarrow$ &
  \tiny PSNR$\uparrow$ / SSIM$\uparrow$ / LPIPS$\downarrow$ \\
\midrule
A) No opacity regularization        & 26.32 / 0.77 / 0.31  & 26.74 / 0.78 / 0.30 \\
B) No scaling regularization        & 29.26 / 0.88 / 0.20  & 29.48 / 0.88 / 0.19 \\
C) No sample rounding               & 28.85 / 0.87 / 0.22  & 29.44 / 0.88 / 0.20 \\
D) No defensive sampling            & 29.20 / 0.88 / 0.20  & 29.45 / 0.88 / 0.19 \\
E) No custom gradient (autodiff)    & 22.35 / 0.62 / 0.51  & 23.69 / 0.64 / 0.47 \\
F) Ours (full model)               & 29.29 / 0.88 / 0.20 & 29.52 / 0.88 / 0.19 \\
\bottomrule
\end{tabularx}

\caption{We ablate our model on six scenes (three outdoor scenes and three indoor scenes) from mip-NeRF 360. We measure the effects of opacity and scaling regularization (A and B), sample rounding (C), defensive sampling (D) and our custom gradient estimator (E). We also quantify the effects of refinement by reporting all values before and after 5K refinement iterations.}
\label{tab:ablations}
\vspace{-10px}
\end{table}

\section{Discussion \& Future Work}
\label{sec:discussion}

Eulerian Gaussian Splatting (EGS) introduces a probabilistic framework for fitting a collection of Gaussians that faithfully represent a set of multiview images. Unlike prior 3DGS-based approaches, our method relies solely on gradient cues and requires no heuristics to prevent convergence to poor local minima.

We introduce Hashed Probability Pyramids, a compact yet expressive representation for high-resolution continuous probability distributions, and a control variate gradient estimator that stabilizes training despite the inherent variance of our probabilistic formulation. Together, these components enable reliable and heuristic-free optimization.

While our approach marks a significant step toward more robust and principled optimization, it remains slower and more memory-intensive than standard 3DGS due to the large number of samples and hash-grid queries required during training. Future work could explore more efficient sampling and variance reduction strategies to further improve training speed and memory efficiency, bringing probabilistic splatting closer to real-time practicality.

In summary, EGS offers a principled bridge between continuous-field optimization and discrete Gaussian rendering, demonstrating that sampling-based formulations can achieve high-quality reconstructions without heuristic intervention. By using an optimizable probabilistic model for 3D Gaussians, our framework combines the efficiency of standard splatting with the robustness of continuous fields.

\paragraph{Acknowledgements.} This work was supported in part by the NVIDIA Academic Grant Program and NSF cooperative agreement PHY\-2019786 (an NSF AI Institute, iaifi.org).

\newpage

{\small
\bibliographystyle{ieeenat_fullname}
\bibliography{11_references}
}

\ifarxiv \clearpage \onecolumn \setcounter{page}{1} \setcounter{figure}{0} \appendix \suptitle{Eulerian Gaussian Splatting using Hashed Probability Pyramids}

\label{sec:appendix_section}

\listofsuppcontents

\supsection{Hashed Probability Pyramids}

\supsubsection{Hashing Function}

We use the same hash encoding function as Instant-NGP~\cite{ingp} to map samples from a bin at level $\ell$ to a $2\times 2\times 2$ block of level $\ell+1$.  For a level $\ell$ with resolution $N_\ell$ satisfying $N_\ell^3 \leq 8B$, the mapping $T_\ell$ is the identity mapping, meaning that the bin index $\volindex$ gets mapped to the $2\times 2\times 2$ distribution at the index $\volindex$. For levels $\ell$ with $N_\ell^3 > 8B$, we set:
\begin{equation}
T(i, j, k) =  \left( i  \cdot \pi_1  \oplus j \cdot \pi_2 \oplus  k \cdot \pi_3 \right) \bmod B,
\end{equation}
where $\volindex = (i, j, k)$ is the 3D index, $\pi_1, \pi_2, \pi_3$ are three large prime numbers (chosen to be the same as~\cite{ingp}), and $\oplus$ denotes the bitwise XOR operation. Note that even though all levels share the same $\pi_1, \pi_2, \pi_3$ the geometric scaling of the grid resolution $N_\ell$ effectively decorrelates the hash collisions across the hierarchy. This is because the same $i,j,k$ indices correspond to radically different spatial coordinates in different levels.

A hashed probability pyramid with the experimental settings described in \cref{sec:probpyramidsettings} uses 86 million parameters to represent a $4096^3$ grid. This requires 0.33 GB of memory, which is smaller than the memory required to store a dense $4096^3$ grid (275 GB) by a factor of more than $800$. 

\supsubsection{Probability Grid Normalization}


In order to sample from our hashed probability pyramid we need to normalize all per-level distributions. More specifically, we need a procedure for creating grids of (normalized) distributions of resolution $R\times R\times R$. For the coarsest distribution $\ell=0$, we have $R=N_0$, and for all other levels $\ell > 1, ..., L-1$, $R = 2$. Given a grid of $R\times R\times R$ parameters $\theta_\nu$ where $\nu\in\{0, ..., R-1\}^3$, we do this by defining:
\begin{equation}
    p(\mu)= R^3\cdot\frac{\sum_{\nu }\exp({\theta_\nu})\mathds{1}\left[\lfloor\mu R\rfloor=\nu\right]}{\sum_{\volindex'} \exp(\theta_{\nu'})},
\end{equation}
for all $\mu \in [0, 1]^3$. The factor $R^3$ is needed to satisfy $\int p(\mu)d\mu = 1$. Here, each $p(\mu)$ is an $R\times R\times R$ block of $f^{(\ell)}_\theta(\mu)$.

Note that this normalization operation must be computed at most $B$ times per level. This is because for levels $\ell$ with parameter sharing (\ie, levels $\ell$ such that $N_\ell^3 > 8B$) there are only $B$ distributions, and for levels without parameter sharing there are fewer than $B$ distributions.



\supsubsection{Additional Remarks}

In addition to making high-resolution probability grids tractable, hashed probability pyramids impose a natural coarse-to-fine hierarchy that aids learning. Spatially nearby samples share coarse-level blocks before diverging at finer resolutions, and this promotes large-scale smoothness while preserving local detail. Importantly, the model automatically automatically allocates capacity across scales through end-to-end optimization, without manually-designed heuristics or human intervention.

Another notable property of our sampling-based approach is that the gradient updates naturally drive down the probabilities of Gaussians that do not contribute to any of the rendered images. Specifically, since the hashed probability pyramid is normalized globally to integrate to one, each time the model drives the probability up at a bin, it also reduces the probabilities everywhere else. This includes reducing the probability in bins with Gaussian samples that are occluded or outside any of the camera views, never contributing to the gradient directly. This in essence acts as an implicit regularizer during training.

\supsection{Gradient Estimator}

\supsubsection{Estimator Intuition}

In \cref{sec:estimator} of the main paper, we describe various gradient estimators, and in \cref{sec:ablations} we show through ablations that our unbiased control variate gradient estimator outperforms the pathwise estimator provided by autodiff. Here we elaborate on these findings. We first provide mathematical intuition using a simplified 1D additive rendering model. Then, in \cref{sec:gradient_visualizations}, we further support these conclusions with an empirical analysis of the gradients produced by each method for the full 3D problem. 

Our simplified 1D rendering model replaces  alpha compositing with summation. We write it as
\begin{equation}
    F(\mu) =  \sum_{i=0}^{M-1} f( \mu_i),
\end{equation}
where $f(\cdot)$ is the rendering of the $i$th Gaussian whose center is $\mu_i$.

As in the main paper, our rendering model is probabilistic:
\begin{equation}
\begin{aligned}
I = \mathbb{E}_{\mu\sim p_\theta(\mu)}\left[F (\mu)\right],
\end{aligned}
\end{equation}
where $p_\theta(\mu)$ is the joint probability distribution over a collection of $M$ Gaussian centers in one dimension, $\mu=(\mu_0, ..., \mu_{M-1}), \mu_i\in\mathbb{R}$.

Our goal is to obtain an unbiased estimator for the gradient of this rendering model $\nabla_\theta I$ with variance that is as low as possible. One common unbiased estimator used in machine learning and computer graphics is the score-based estimator:
\begin{equation}
\begin{aligned} \label{eq:scoreestjoint}
g_j &\overset{\Delta}{=} \nabla_\theta \mathbb{E}_{\mu\sim p_\theta(\mu)}\left[F(\mu)\right] \\
&= \nabla_\theta \int p_\theta(\mu) F(\mu) \; d\mu\\
&=  \int p_\theta(\mu) F(\mu) \nabla_\theta \log p_\theta(\mu) \; d\mu \\ 
&= \mathbb{E}_{\mu\sim p_\theta(\mu)} \left[\sum_{i=0}^{M-1} F(\mu) \nabla_\theta \log p_\theta(\mu_i) \right].
\end{aligned}
\end{equation}
Here, we used the independence of samples, $\log p_\theta(\mu) = \sum_{i=0}^{M-1}\log p_\theta(\mu_i)$.

Alternatively, we can use the linearity of our additive rendering model to derive a score-based estimator using the marginal distribution over a single Gaussian instead of the joint $M$-Gaussian distribution:
\begin{equation}
\begin{aligned} \label{eq:scoreest1}
g_m &\overset{\Delta}{=}\nabla_\theta \mathbb{E}_{\mu\sim p_\theta(\mu)}\left[ \sum_{i=0}^{M-1} f(\mu_i)\right] \\
&= \sum_{i=0}^{M-1} \nabla_\theta\mathbb{E}_{\mu_i\sim p_\theta(\mu)}[f(\mu_i)]\\
&= \sum_{i=0}^{M-1} \mathbb{E}_{\mu_i\sim p_\theta(\mu)}[f(\mu_i) \nabla_\theta\log p_\theta(\mu_i)].
\end{aligned}
\end{equation}

The two estimators in \cref{eq:scoreestjoint} and \cref{eq:scoreest1} look  similar, but their variances are extremely different. The marginal distribution estimator in \cref{eq:scoreest1} provides significantly lower variance than the estimator in \cref{eq:scoreestjoint}. 
To show this analytically, we begin by expanding the expression of the joint estimator:
\begin{equation}
\begin{aligned} \label{eq:scoreestmath}
g_j &= \nabla_\theta \mathbb{E}_{\mu\sim p_\theta(\mu)}\left[F(\mu)\right] \\
&= \mathbb{E}_{\mu\sim p_\theta(\mu)} \left[\sum_{i=0}^{M-1} f(\mu_i) \cdot \sum_{j=0}^{M-1} \nabla_\theta \log p_\theta(\mu_j) \right] \\
&= \mathbb{E}_{\mu\sim p_\theta(\mu)} \left[\sum_{i=0}^{M-1} f(\mu_i) \cdot \nabla_\theta \log p_\theta(\mu_i) + \sum_{i=0}^{M-1} \nabla_\theta \log p(\mu_i) \sum_{j\neq i} f(\mu_j) \right] \\
&=  \underbrace{\mathbb{E}_{\mu\sim p_\theta(\mu)} \left[\sum_{i=0}^{M-1} f(\mu_i) \cdot \nabla_\theta \log p(\mu_i) \right]}_{g_m} +\underbrace{\mathbb{E}_{\mu\sim p_\theta(\mu)} \left[\sum_{i=0}^{M-1} \nabla_\theta \log p(\mu_i) \sum_{j\neq i} f(\mu_j) \right]}_{R}.   \\
\end{aligned}
\end{equation}
Observe that the first term in \cref{eq:scoreestmath} is identical to the marginal estimator $g_m$ in \cref{eq:scoreest1}. The additional ``cross-term'' $R$ has zero mean, and we will show that it adds variance to the entire estimator. Let $s_i = \nabla_\theta \log p_\theta(\mu_i)$ and $a_j = f(\mu_j)$, and define per-sample moments:
\begin{equation}
\begin{aligned}
\mathbb{E}[a_i] &= \mu_a, & \operatorname{Var}(a_i) &= \sigma_a^2, &\mathbb{E}[s_i] &= 0,     & \operatorname{Var}(s_i) &= \sigma_s^2, & \rho = \mathbb{E}[a_is_i].
\end{aligned}
\end{equation}
The cross-term $R$ is $\sum_i\sum_{j\neq i} s_i a_j$ and we can write its variance as:
\begin{equation}
\begin{aligned}
\text{Var}(R) &= \mathbb{E}[R^2] - (\mathbb{E}[R])^2 \\
&= \mathbb{E}\left[\left( \sum_i\sum_{j\neq i}  s_i a_j \right)^2 \right] - 0 \\
&= \sum_i\sum_{j\neq i} \sum_k\sum_{l \neq k} \mathbb{E}[s_i a_j s_k a_l] \\
&= \sigma_s^2 \left(M(M-1) \sigma_a^2 + M (M-1)^2\mu_a^2 \right) + M(M-1)\rho^2.
\end{aligned}
\end{equation}

Notably, the variance of the cross-term is $\mathcal{O}(M^3)$. Likewise, we can derive the variance of the marginal estimator:
\begin{equation}
\begin{aligned}
\text{Var}(g_m) &= \sum_{i=0}^{M-1} \operatorname{Var}(f(\mu_i) \nabla_\theta\log p_\theta(\mu_i)).
\\
&=M\cdot\operatorname{Var}(a_is_i),
\end{aligned}
\end{equation}
where we used the independence $\mu_i$ and $\mu_j$ for $i\neq j$.

This means that the variance of the marginal estimator is $\mathcal{O}(M)$. Because $|\operatorname{Cov}(g_m, R)| \le (\operatorname{Var}(g_m)\operatorname{Var(R)})^{1/2}$, $\operatorname{Cov}(g_m, R)$ is at most $\mathcal{O}(M^2)$. This means that the variance of $R$, which is $\mathcal{O}(M^3)$, dominates the variance for sufficiently large $M$ since $\operatorname{Var}(g_j) = \operatorname{Var}(g_m + R) = \operatorname{Var}(g_m) + \operatorname{Var}(R) + 2 \operatorname{Cov}(g_m, R)$.

The implication is that the variance of the joint $M$-Gaussian gradient estimator scales with the number of Gaussians $M$ much more quickly than the marginal estimator. In contrast, the marginal estimator negates these cross-terms so that updates to a voxel of the probability distribution are only influenced by Gaussians generated by that voxel, which leads to more stable and lower-variance training.

Our estimator in \cref{sec:estimator} of the main paper is inspired by this exact analysis. In this simplified 1D additive problem, the intuitive reason that the joint estimator of \cref{eq:scoreestjoint}  results in much higher variance is that the gradient update of the $i$th sample stemming from $\log p_\theta(\mu_i)$ is multiplied by the entire image $F(\mu)$ instead of just the contribution of the $i$th Gaussian to the image $f(\mu_i)$. This ``mixing'' of contributions across all Gaussians adds variance to the estimator. In the full 3D problem with alpha compositing, there is no analytical equivalent to the marginal estimator, but we can leverage the same intuition by replacing the full image $I$ in the gradient expression with $(I-I_{-i})$, the contribution of the $i$th Gaussian.



\supsubsection{Pathwise Estimator}

As mentioned in the main paper, because our sampling process in \cref{sec:sampling} is differentiable, we can derive a third gradient estimator---the pathwise estimator---using the ``reparameterization trick''. If we denote our sampling process by $\mu_i = g(u_i, \theta)$, which maps $u_i \sim \textrm{U}([0, 1])$ to $\mu_i\sim p_\theta$, we can write the pathwise estimator as
\begin{equation}
    \begin{aligned}
        \nabla_\theta F(\mu) =  \mathbb{E}_{u \sim \mathrm{U}([0,1])}\!\left[ \sum_{i=0}^{M-1} \nabla_{\mu_i} F(\mu)  \nabla_\theta g(u_i, \theta) \right].
    \end{aligned}
\end{equation}
This is another unbiased estimator, and it only requires backpropagation through the renderer to evaluate $\nabla_{\mu_i} F(\mu)$.

In the next section, we include a comparison of the variances of the three estimators for our 1D problem, showing that the marginal score-based estimator has the lowest variance. We then show that the analogous estimator of the original 3D problem, which we introduced in \cref{sec:estimator} of the original paper, has lower variance than the alternative estimators in 3D empirical studies.

\supsubsection{Empirical Evidence for Variance Reduction}\label{sec:gradient_visualizations}

We showed through ablations in \cref{sec:ablations} of the main paper that our model trains much better with our control variate estimator than with standard auto differentiation gradients (\ie, the pathwise estimator). Here we provide additional qualitative comparisons that further support this finding.

In 1D, we construct a ground-truth signal by summing 15 Gaussians with randomly sampled widths, amplitudes, and positions, as shown on the left of \cref{fig:1d_comparison}. We initialize a uniform probability distribution representing the likelihood of placing a Gaussian center at each position along the signal. The middle plot shows the gradient of this probability distribution. The rightmost plot compares gradient variance across the three estimators: the marginal estimator, the joint $M$-Gaussian estimator, and the pathwise estimator. Among these, the marginal estimator exhibits the lowest overall variance (particularly in empty regions of the signal) indicating that probabilities in these areas decrease reliably. In contrast, the joint $M$-Gaussian estimator has the highest variance.

\begin{suppfigure}[h]
    \centering
\includegraphics[width=0.9\linewidth]{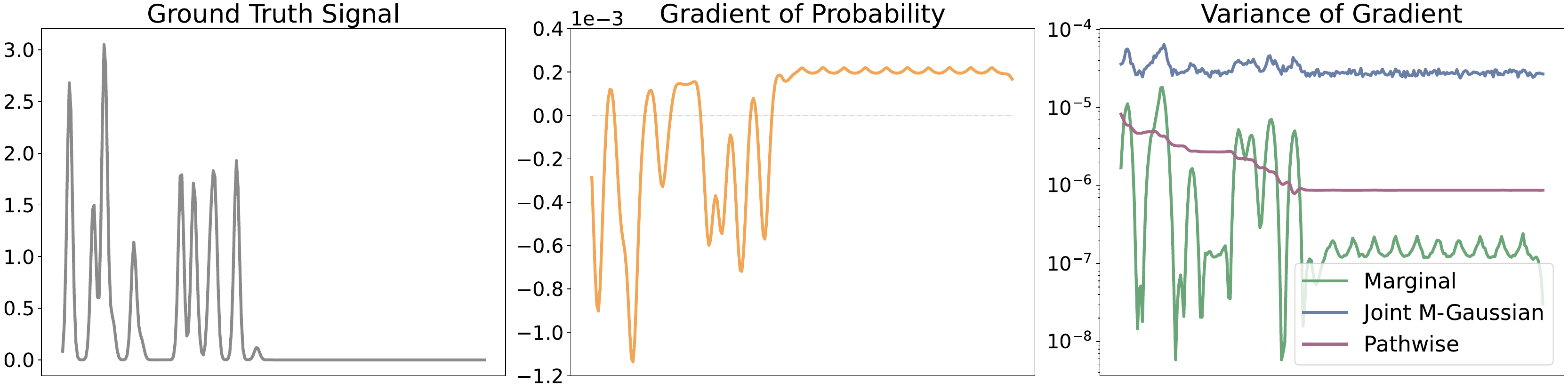}
    \caption{1D illustration of the three gradient estimators. The ground-truth signal (left) is constructed by summing 15 Gaussians with random widths, amplitudes, and positions. We initialize a uniform probability distribution over possible Gaussian center locations and visualize its gradient (middle). The gradient variance (right) is compared across three estimators: marginal, joint $M$-Gaussian, and pathwise. The marginal estimator exhibits the lowest overall variance, while the joint $M$-Gaussian estimator has the highest. \label{fig:1d_comparison}}
\label{fig:variance_comparison}
\end{suppfigure}

In 3D, we visualize the estimated gradients of our probability distribution for three estimators---pathwise, standard score function, and our proposed control variate, which is inspired by the marginal estimator from the 1D experiments. We define a single level, dense $128^3$ probability grid (\ie,  $L=1$, $N_0 = 128$, and $B>2^{18}$) initialized uniformly. Using a single image from the bicycle scene in the mip-NeRF~360 dataset~\cite{mipnerf360}, downsampled by a factor of 16 (4 times smaller than standard training resolution), we compute 100 independent gradient estimates of the loss for each method, with each estimate using one million samples. We then visualize the mean and variance across these 100 gradient estimates.

\begin{suppfigure}[h]
    \centering
\includegraphics[width=1.0\linewidth]{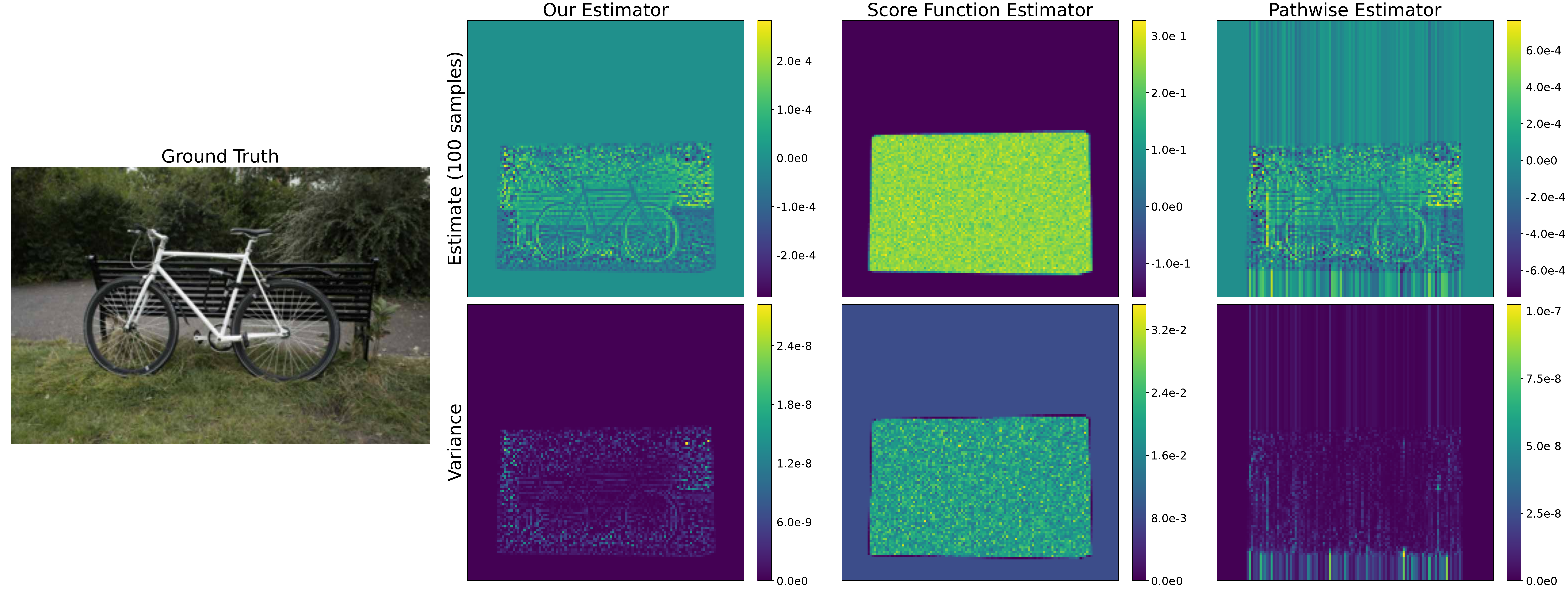}
    \caption{3D gradient variance comparison across estimators. We visualize the mean (top) and variance (bottom) of 100 independently estimated probability gradients for the bicycle scene from mip-NeRF~360~\cite{mipnerf360}. Our control variate estimator exhibits substantially lower variance than the pathwise and standard score function estimators, particularly in empty regions of space, leading to more stable and efficient training.}
\label{fig:variance_comparison}
\end{suppfigure}

Each column of Fig.~\ref{fig:variance_comparison} shows the estimated gradient (top) and its variance (bottom) for one of the three estimators. Our control variate estimator achieves several orders of magnitude lower variance than both the pathwise and score function estimators. Visually, it maintains low variance even in regions corresponding to Gaussians outside the image plane, similar to the behavior of the marginal estimator in 1D. This is a desirable property that allows the model to reliably reduce probability in empty regions and avoid wasted capacity. In contrast, the pathwise estimator exhibits structured, striation-like variance patterns that we find detrimental to training stability in practice.

\supsubsection{Gradient Estimator Derivation}\label{sec:simplifying_difference}

A key to making our gradient estimator tractable is that the difference $(I - I_{-i})$ between the full rendering of an image $I$ and its rendering $I_{-i}$ with the $i$th Gaussian removed simplifies to known quantities that are already being computed during the backward pass to update the attribute grids. The derivation follows.

The color of a pixel is determined by the Gaussians that intersect the ray that backprojects from that pixel. Let these $K$ Gaussians have colors $c_1, ..., c_K \in \mathbb{R}^3$ and opacities
$o_1, ..., o_K \in [0, 1]$. Then the value of the pixel is
\begin{equation} \label{eq:alphacomp}
    I = \sum_{k=1}^K c_k\alpha_k T_k,
\end{equation}
where the transmittance is defined as
\begin{equation}
    T_k = \prod_{k'=1}^{k-1}
    (1-\alpha_{k'}),
\end{equation}
and $\alpha_k$ is the pixel opacity determined by $o_k$ and the pixel position. 

We split the color from \cref{eq:alphacomp} into three terms, given any Gaussian index $i \in \{1, ..., K\}$:
\begin{equation} \label{eq:img}
    I = \sum_{k=1}^{i-1} c_k \alpha_k T_k + c_i \alpha_i T_i + \sum_{k={i+1}}^K c_k\alpha_k T_k,
\end{equation}
and we observe that the color obtained by omitting the $i$th Gaussian is
\begin{equation} \label{eq:imgsub}
    I_{-i} = \sum_{k=1}^{i-1} c_k \alpha_k T_k + \frac{1}{1-\alpha_i}\sum_{k={i+1}}^K c_k\alpha_kT_k,
\end{equation}
where the division by $1-\alpha_i$ removes the effect of the $i$th Gaussian from the transmittance of Gaussians with indices $k > i$. Subtracting \cref{eq:imgsub} from \cref{eq:img} yields
\begin{equation} \label{eq:imgdiffsupp}
    I - I_{-i} = c_i\alpha_i T_i- \frac{\alpha_i}{1-\alpha_i}\sum_{k={i+1}}^K c_k\alpha_kT_k. 
\end{equation}

On the other hand, taking the derivative of $I$ from \cref{eq:img} with respect to $\alpha_i$, we obtain
\begin{equation}
    \frac{\partial I}{\partial \alpha_i} = c_i T_i + \frac{1}{1-\alpha_i} \sum_{k=i+1}^K c_k \alpha_k  T_k,
\end{equation}
which when combined with \cref{eq:imgdiffsupp} yields
\begin{equation}
    I - I_{-i} = \alpha_i \frac{\partial I}{\partial \alpha_i}.
\end{equation}

Additionally, in the 3DGS render operation, the pixel opacity $\alpha_i$  is proportional to the Gaussian opacity $o_i$, so
\begin{equation}
    \alpha_i\frac{\partial I}{\partial \alpha_i} = o_i\frac{\partial I}{\partial o_i},
\end{equation}
and we can write the difference as
\begin{equation}\label{eq:difference}
I - I_{-i} = o_i \frac{\partial I(x)}{\partial o_i}.
\end{equation}
Thus, computing the difference $(I-I_{-i})$ does not require an additional forward pass to render $I_{-i}$. Instead, it can be computed directly from the opacity $o_i$ and its gradient $\partial \mathcal{L}/\partial o_i$, both of which are readily available from the standard rendering process. By substituting \cref{eq:difference} into \cref{eq:leave-one-out} from the main paper, we obtain the final expression for our gradient estimator:
\begin{equation}
\nabla_\theta I
=\mathbb{E}_{\mu\sim p_\theta(\mu)}\left[
\sum_{i=0}^{M-1}o_i \frac{\partial I}{\partial o_i} \cdot\nabla_\theta\log p_\theta(\mu_i)\right].
\end{equation}

\supsection{Implementation Details}

\supsubsection{Hash Encoding Grid Settings.}

\begin{suppfigure}[h]
\centering
\begin{lstlisting}[label={lst:hashgrid}]
encoding_config = {
    'otype': 'HashGrid',
    'n_levels': 13,
    'n_features_per_level': 1 if opacity else 8,
    'log2_hashmap_size': 23,
    'base_resolution': 2,
    'per_level_scale': 2.0,
    'interpolation': 'Smoothstep'
}
opacity_network_config = {
    'otype': 'FullyFusedMLP',
    'activation': 'LeakyReLU',
    'output_activation': 'None',
    'n_neurons': 32,
    'n_hidden_layers': 1
}
color_network_config = {
    'otype': 'FullyFusedMLP',
    'output_activation': 'None',
    'n_hidden_layers': 0
}
scaling&rotation_network_config = {
    'otype': 'FullyFusedMLP',
    'activation': 'LeakyReLU',
    'output_activation': 'None',
    'n_neurons': 32,
    'n_hidden_layers': 1
}
\end{lstlisting}
\caption{\label{fig:hashgrid} Hash grid configuration for the \texttt{tiny-cuda-nn} implementation~\cite{tiny-cuda-nn}}
\end{suppfigure}

We use small MLPs with a single hidden layer of size 32 to map the opacity and scale/rotation gridded features to output attributes, and we map color features with a single linear layer. \Cref{fig:hashgrid} shows the exact hash encoding configurations.

Our parameterization of scale $s\in\mathbb{R}^3$ and opacity $o \in [0, 1]$ is defined as:
\begin{align}
    s = \textrm{softplus}(\tilde{s} + \textrm{softplus}^{-1}(s_0)), \\
    o = \textrm{sigmoid}(\tilde{o} + \textrm{sigmoid}^{-1}(o_0)),
\end{align}
where $\tilde{o}$ and $\tilde{s}$ are the outputs of the hash encoding, and $o_0$ and $s_o$ are the initial values for opacity and scale. This relies on the hash encoding output being small at the start of training, which in practice is $\sim 10^{-6}$. We initialize our model with $o_0 = 0.05$ and $s_0=0.0006$. 

We represent the three color channels as degree three spherical harmonics. To prevent higher-order coefficients from dominating during early training, we multiply the color coefficients by a factor that depends on their degree $\ell$, such that the color is:
\begin{equation}
    C(\boldsymbol{\omega})=\sum_{\ell=0}^3\sum_{m=-\ell}^{\ell} \alpha^\ell \cdot c_\ell^m Y_\ell^m(\boldsymbol{\omega}),
\end{equation}
where $\boldsymbol{\omega}$ is the view direction and $\alpha=0.2$, and $c_\ell^m$ are the optimizable grid parameters.

\supsubsection{Custom Gradient Implementation}
The pathwise (auto-differentiation) gradient of the loss can be written as the sum over Gaussian centers $\mu_i$:
\begin{equation}
\begin{aligned}
\nabla_\theta \mathcal{L}(I, I^\text{GT}) =\sum_{i=1}^{M-1} \frac{\partial \mathcal{L}}{\partial \mu_i} \cdot 
        \frac{\partial \mu_i}{\partial \theta}.
\end{aligned}
\end{equation}

By manipulating the gradients of these terms, we can force the model to use our control variate gradient estimator during the backward pass. We replace the sampled positions $\mu_i$ with a surrogate variable whose gradient tracks the log-probability:
\begin{equation}
\mu_i \;\;\longrightarrow\;\; 
\stopgrad(\mu_i) + \log p_\theta(\mu_i) - \stopgrad\!\big(\log p_\theta(\mu_i)\big),
\end{equation}
where $\stopgrad(\cdot)$ denotes the stop-gradient operator.  
This ensures that gradients flow through $\log \p_\theta(\mu_i)$ while the value of $\mu_i$ is preserved in the forward pass.

We also modify the upstream gradient $\partial \mathcal{L} / \partial \mu_i$, by substituting
\begin{equation}
\frac{\partial \mathcal{L}}{\partial \mu_i}
\;\;\longrightarrow\;\;
o_i\,\frac{\partial \mathcal{L}}{\partial o_i}
\end{equation}
inside the renderer backward pass.  

We implement these changes in practice by augmenting $\bmu$ and $\partial I(x; \bmu)/\partial \bmu$ with an additional channel to store the surrogate gradients and zero out the gradients elsewhere. These changes enable us to use our custom gradient estimator while preserving the efficiency of the standard rendering pipeline.

\supsection{Additional Training Details and Results}

\supsubsection{Renderer modifications}

Our rendering pipeline utilizes an adapted version of the Slang.D-based 3DGS rasterizer ~\cite{kopanas2025slanggs,bangaru2023slangd}, tailored to accommodate our custom gradient estimator.

Additionally, we cap the number of rendered Gaussians at 7.5 million per frame by randomly selecting 7.5 million in-frustum Gaussians to keep if more than 7.5 million Gaussian samples are within the view frustum.

\supsubsection{Refinement details}

During the $5$K refinement steps that we apply to the set of Gaussians sampled from our model, we deviate from standard optimization in a few ways. First, we continue frustum-culling the Gaussians during refinement in order to remain consistent with the sampling model. Second, we fix the positions of all Gaussians and we use a reduced learning rate of $5\cdot 10^{-3}$ for opacity. We set all other learning rates to be consistent with the learning rates used by 3DGS at the start of training~\cite{3DGS}. Finally, we disable all adaptive density control modules (pruning, cloning, culling) so that the set of Gaussians and their positions are fixed.

\supsubsection{Near Plane Culling}

The standard 3DGS renderer employs a $z$-threshold to discard Gaussians within a small margin of the camera's image plane, mitigating artifacts caused by ``floater'' artifacts. We retain the default $z=0.2$ for all datasets except Tanks and Temples~\cite{TANT}, where we lower it to $z=0.05$ to account for objects being positioned closer to the camera.

\supsubsection{Additional Qualitative Comparisons}

Here we provide a few complementary visualizations for our ablation studies in~\cref{sec:ablations}. \cref{fig:opacity_ablation} shows a comparison of our full model with a variant trained without opacity regularization. The absence of opacity regularization leads to regions dominated by large, high-opacity Gaussians, resulting in poor reconstruction of fine textures such as distant grass. In contrast, opacity regularization enables better preservation of detail. \cref{fig:estimator_ablation} shows a comparison of our approach with a model trained using the pathwise gradient estimator. Although both estimators are unbiased, our proposed estimator more effectively optimizes the underlying probability distribution, yielding a more accurate scene reconstruction.

\begin{suppfigure}[H]
    \centering
\includegraphics[width=1.0\linewidth]{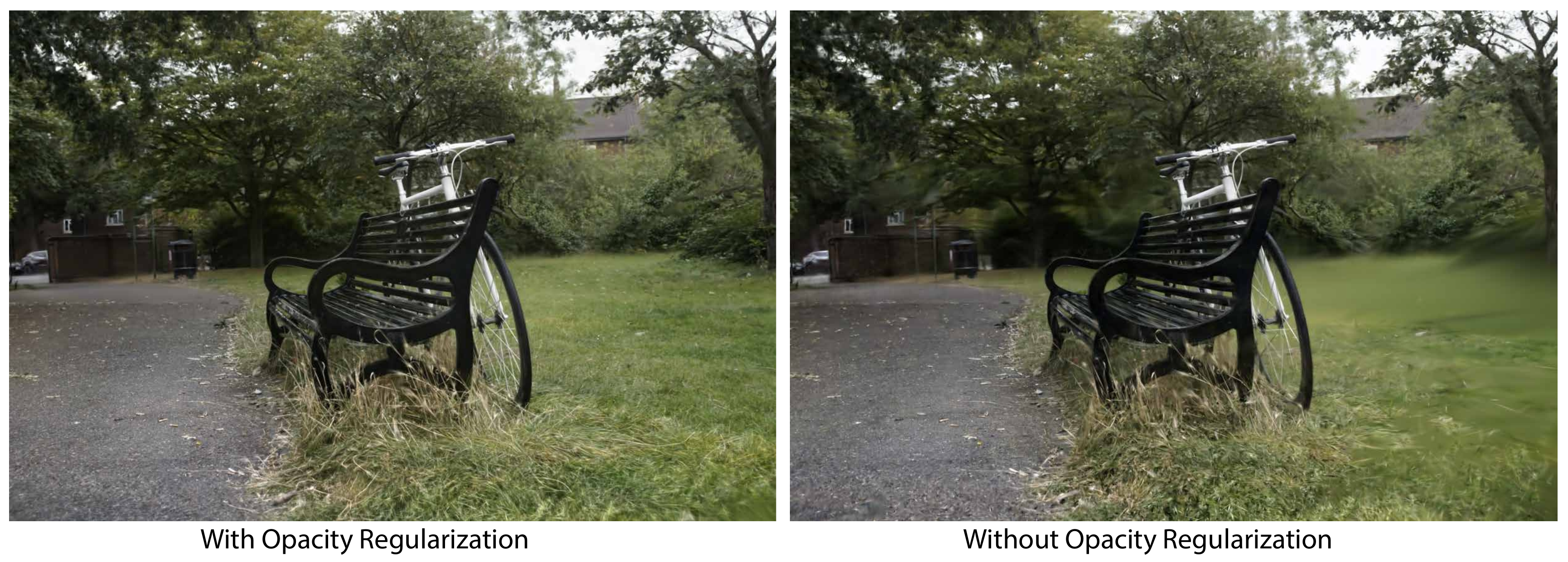}
    \caption{The results of our method with the opacity regularization (left) and without it (right). Opacity regularization prevents large high-opacity Gaussians from dominating regions such as distant grass, and enables preserves fine detail. \label{fig:opacity_ablation}}
\end{suppfigure}

\begin{suppfigure}[H]
    \centering
\includegraphics[width=1.0\linewidth]{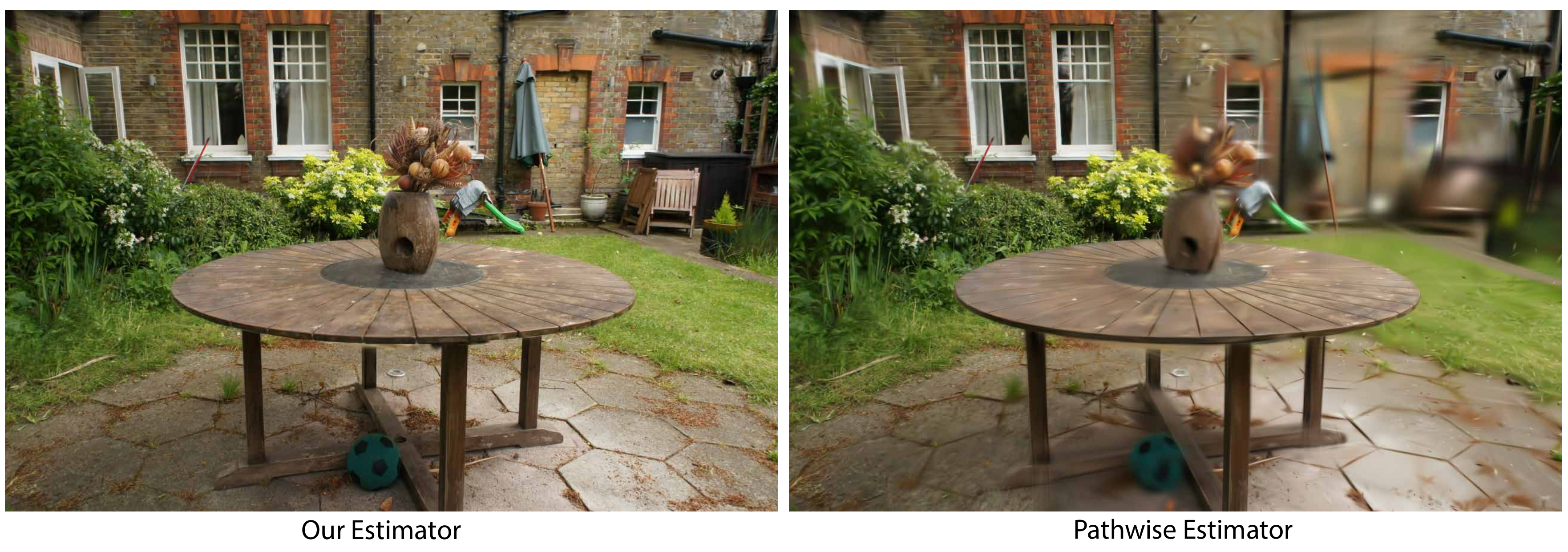}
    \caption{Our control variate estimator (left) produces sharper, more accurate reconstructions than the pathwise gradient estimator (right), which fails to optimize the probability distribution effectively. \label{fig:estimator_ablation}}
\end{suppfigure}

\supsubsection{Initialization with COLMAP}

We also explored initializing the weights of the hashed probability pyramid using the points output from COLMAP. Specifically, we set the probability value of occupied voxels to be twice that of empty ones. This initialization reduces the need for the model to ``discover'' these regions during optimization. Empirically, we find that this leads to faster convergence. We evaluate this strategy on four randomly selected scenes from mip-NeRF 360 (two indoor and two outdoor scenes) and observe that COLMAP-based initialization achieves the same PSNR approximately $15\%$ faster compared to uniform initialization.

\supsubsection{Dependence on the number of samples}

Our experiments indicate that increasing the number of Gaussian samples improves reconstruction quality, at the cost of moderately increased training time. We provide a quantitative comparison in \cref{tab:numgauss}, where we vary the number of samples on the mip-NeRF 360 ``Room'' scene.

As shown in the table, using more samples consistently improves PSNR and LPIPS, both before and after refinement, while SSIM remains largely stable. However, these gains come with a gradual increase in training time. This highlights a trade-off between reconstruction fidelity and computational cost, where larger sample counts yield better performance but incur higher training overhead.

\begin{table}[h]
\centering
\footnotesize
\setlength{\tabcolsep}{3pt}
\begin{tabularx}{0.6\linewidth}{@{}l *{3}{Y} @{}}
\toprule
\multicolumn{1}{c}{\textbf{\# Samples}} &
\multicolumn{1}{c}{\textbf{Pre-refinement}} &
\multicolumn{1}{c}{\textbf{Post-refinement}} &
\multicolumn{1}{c}{\textbf{Training Time}} \\
& \tiny PSNR$\uparrow$ / SSIM$\uparrow$ / LPIPS$\downarrow$ & \tiny PSNR$\uparrow$ / SSIM$\uparrow$ / LPIPS$\downarrow$ &  \tiny (in hours) \vspace{1mm}
\\
A) $1.5 \cdot10^{7}$ & 32.09 / 0.93 / 0.26 & 32.41 / 0.93 / 0.25 & 5.87\\
B) $10^{7}$ & 32.04 / 0.93 / 0.27 & 32.38 / 0.93 / 0.26 &  5.22\\
C) $5 \cdot10^{6}$  & 31.71 / 0.93 / 0.28 & 31.90 / 0.93 / 0.27 & 4.70\\
D) $2.5 \cdot10^{6}$  & 31.51 / 0.92 / 0.29 & 31.70 / 0.92 / 0.29  & 4.23\\
\bottomrule
\end{tabularx}
\caption{Impact on model performance (quality and training time) of the number of sampled Gaussians.}
\label{tab:numgauss}
\end{table} \fi




\end{document}